\documentclass[journal]{IEEEtran}
\usepackage{threeparttable,booktabs}
 \usepackage{mathrsfs}
 
\usepackage{amsmath}
\usepackage{amssymb}
\usepackage{graphicx}

\usepackage{subfigure}

\usepackage{algorithm}
\usepackage{algorithmicx}
\usepackage{algpseudocode}
\usepackage{amsmath}
\usepackage{wrapfig}
\usepackage{graphicx}
\usepackage[justification=centering]{caption}

\usepackage{ifpdf}

\ifCLASSINFOpdf
\else
\fi
\usepackage{amsmath}
\usepackage{array}
\begin{document}
%
\title{Locally Linear Region Knowledge Distillation}
%
%
%

\author{Xiang Deng and Zhongfei (Mark) Zhang,
\thanks{Xiang Deng is with the Department
of Computer Science, State University of New York at Binghamton, e-mail: xdeng7@binghamton.edu.}
\thanks{Zhongfei (Mark) Zhang is with the Department of Computer Science, State University of New York at Binghamton, e-mail: zhongfei@cs.binghamton.edu.}
}

\maketitle

\begin{abstract}
Knowledge distillation (KD) is an effective technique to transfer knowledge from one neural network (teacher) to another (student), thus improving the performance of the student.
To make the student better mimic the behavior of the teacher,
the existing work focuses on designing different criteria to align their logits or representations.
Different from these efforts, we address knowledge distillation from a novel data perspective.
We argue that transferring knowledge at sparse training data points cannot enable the student to well capture the local shape of the teacher function.
To address this issue, we propose locally linear region knowledge distillation ($\rm L^2$RKD) which transfers the knowledge in local, linear regions from a teacher to a student.
This is achieved by enforcing the student to mimic the outputs of the teacher function in local, linear regions.
To the end, the student is able to better capture the local shape of the teacher function and thus achieves a better performance.
Despite its simplicity, extensive experiments demonstrate that $\rm L^2$RKD is superior to the original KD in many aspects as it outperforms KD and the other state-of-the-art approaches by a large margin, shows robustness and superiority under few-shot settings, and is more compatible with the existing distillation approaches to further improve their performances significantly.
\end{abstract}


%
\IEEEpeerreviewmaketitle

\section{Introduction}
Deep neural networks (DNNs) have achieved remarkable performances on a wide range of applications in artificial intelligence including computer vision \cite{krizhevsky2012imagenet, girshick2015fast, zhang2020sg} and natural language processing \cite{andreas2016learning, serban2016generating}.
The performance gain comes at a high cost of computation and memory in inference.
This severely restricts their implementation on resource-limited hardware such as phones, watches, and robots.
One solution to this problem is knowledge distillation which transfers the knowledge from a large network (teacher) to a small and fast one (student).
To the end, the student obtains a significant performance boost and even possibly matches or surpasses the performance of the teacher.
\par

\begin{figure*}[!t]
     \begin{minipage}{0.32\textwidth}
     \centering
     \includegraphics[height=2.8cm]{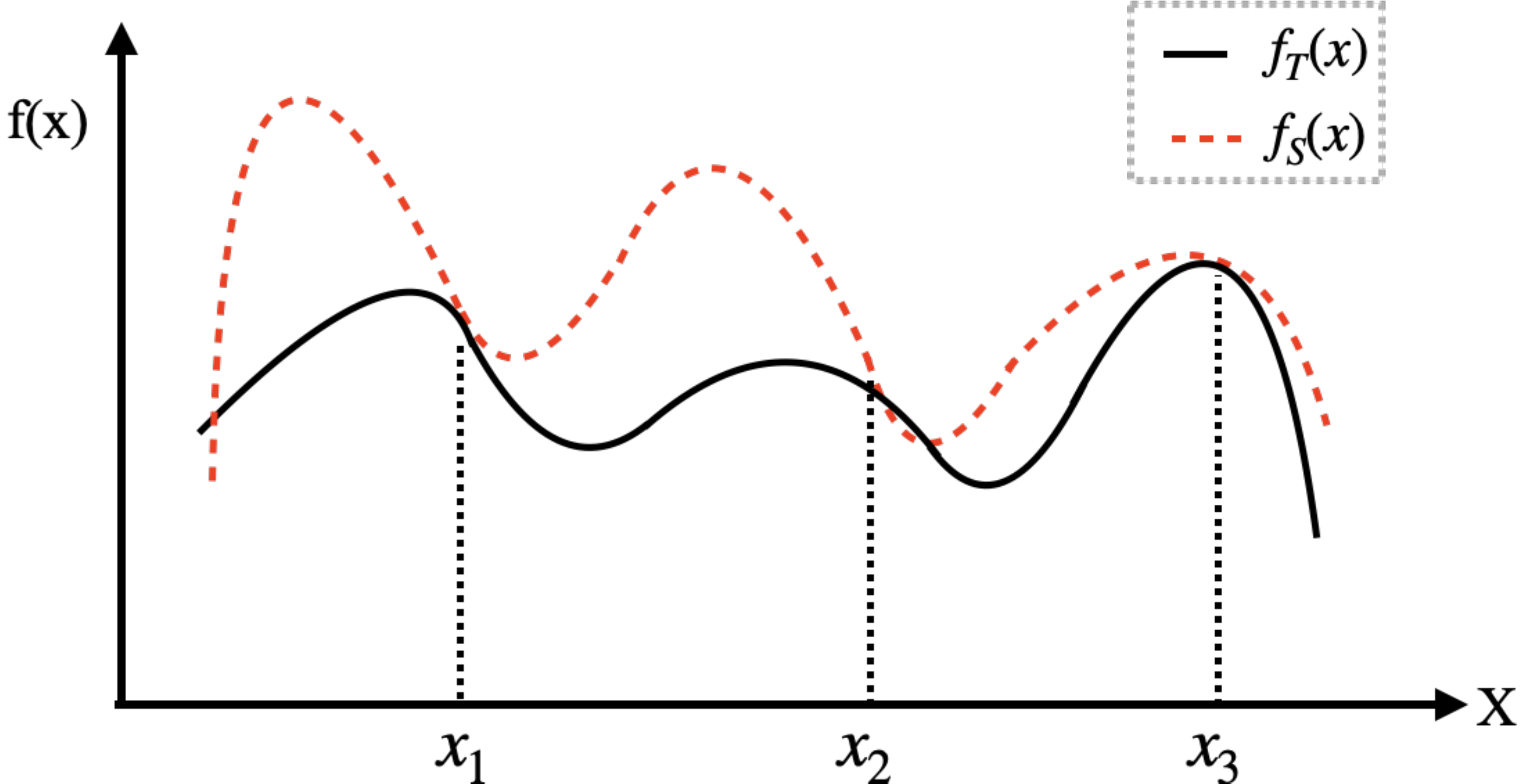}
\setlength{\belowcaptionskip}{-0cm}
     \caption{KD with trainig data}
     \label{f1}
   \end{minipage}\hfill
   \begin{minipage}{0.32\textwidth}
     \centering
     \includegraphics[height=2.8cm]{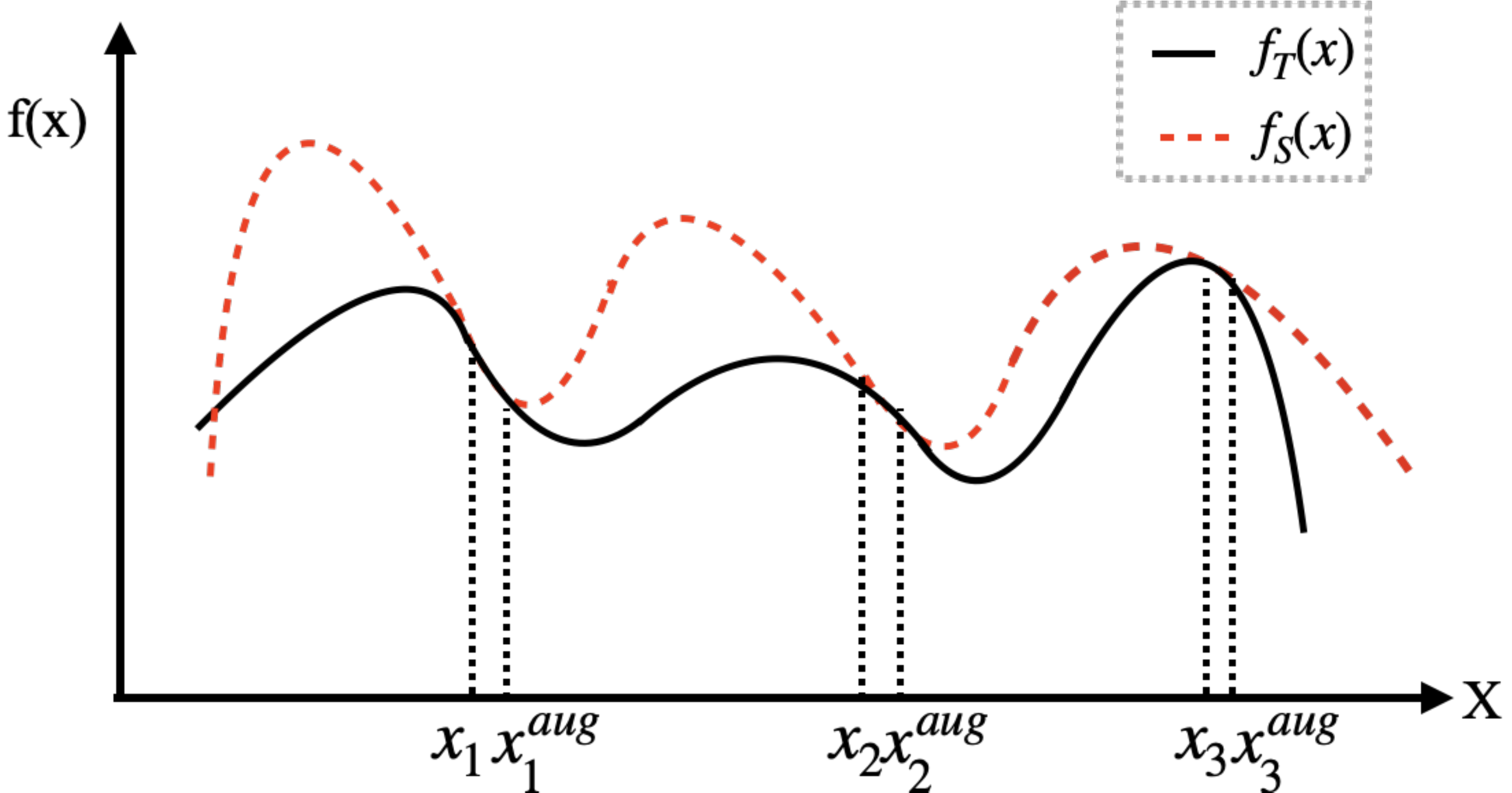}
     \caption{KD with data augmentation}
     \label{f2}
   \end{minipage}\hfill
   \begin{minipage}{0.32\textwidth}
     \centering
     \includegraphics[height=2.8cm]{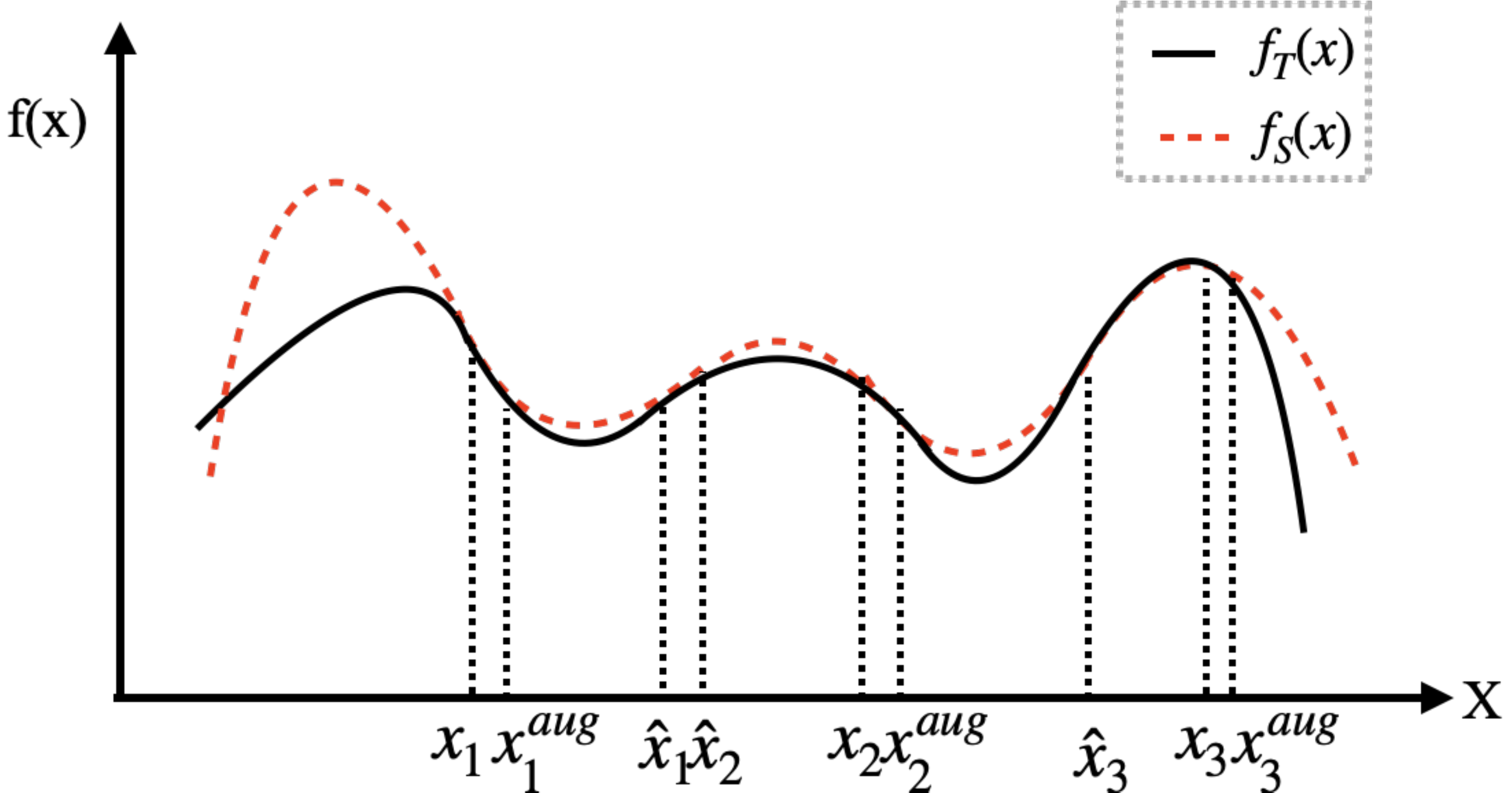}
     \caption{$\rm L^2$RKD}
     \label{f3}
   \end{minipage}  
\end{figure*}

Hinton et al. \cite{hinton2015distilling} proposed the original Knowledge Distillation (KD) that utilizes the soft labels generated by a teacher as the targets to train a student.
From the perspective of function curve fitting, the student function is to learn the local surface shape of the teacher function at each training data point.
It is widely believed that the generalization ability \cite{novak2018sensitivity,kawaguchi2017generalization} of a DNN is highly related to the stability which can be described as the local function shape of a DNN.
When we have a large teacher network which typically has a much better generalization performance than that of the student, we can suppose that the teacher function has a more stable (better) shape within the data distribution than that of the student.
Intuitively, if the student better captures the local shape of the teacher, it can have a better generalization ability.
However, the existing work only focuses on designing different criteria to transfer the knowledge at each individual training data point from a teacher to a student.
Fitting sparse individual data point may not necessarily enable the student to well capture the local shape of the teacher function.
As shown in Figure \ref{f1}, even if the student $f_S$ perfectly fits the teacher $f_T$ at each training data point, i.e., $x_1$, $x_2$, and $x_3$, their local shapes near these data points can be highly different.

To alleviate this problem, the typically used strategy is to train on different but similar data points to the training data.
This is known as data augmentation \cite{simard1998transformation} formalized by the {\em Vicinal Risk Minimization} (VRM) \cite{chapelle2001vicinal} principle.
In VRM, human knowledge is necessary to define a vicinity or neighborhood around each training data point.
Then, additional new data points can be drawn from the vicinity distribution of the training data to enlarge the support of the training data distribution.
For example, in image classification, it is common to define the vicinity of an image as the set of its random flippings and crops after a mild padding.
Nevertheless, data augmentation has its own limitation that a new generated data point is very close to the original training data point, since they contain almost the identical objective only with different backgrounds caused by padding or cropping.
Due to this limitation, as shown in Figure \ref{f2}, even if the student $f_S$ fits the teacher $f_T$ at all the training data points (i.e., $x_1$, $x_2$, and $x_3$) and the augmented data points (i.e., $x_1^{aug}$, $x_2^{aug}$, and $x_3^{aug}$), their local shapes can still differ substantially.
\par

On the other hand, we observe that the existing studies ignore an advantage of knowledge distillation, i.e., for any given input sample, the teacher can always provide the soft label for the input sample even if the sample is out-of-distribution.
It is obvious that the soft labels for out-of-distribution samples are semantically meaningless.
However, if these out-of-distribution samples are not far from the real data distribution, distilling on these samples may be able to assist the student to better capture the local, in-distribution shape of the teacher.
In light of this, we propose locally linear region knowledge distillation ($\rm L^2$RKD) which enforces the student to mimic the shape of the teacher function in local, linear regions.
As shown in Figure \ref{f3}, instead of only fitting the data points near the training samples, $\rm L^2$RKD also enforces the student to fit the data points in the linear region between two training or augmented samples, e.g., $\hat x_1$ and $\hat x_2$ in the linear region between $x_1^{aug}$ and $x_2^{aug}$, and $\hat x_3$ in the linear region between $x_2^{aug}$ and $x_3^{aug}$.
Each sample can be considered as a high-dimensional vector or point.
Two points can determine a linear region so that we can easily draw points from the linear region by using linear algebra.
As seen from Figure 3, distilling knowledge in local, linear regions can make the student better capture the local shape of the teacher function.
We empirically verify (in Section \ref{shape}) that the teacher-student shape differences of $\rm L^2$RKD are significantly smaller that those of KD on real datasets, thus leading to a better performance.

\par

To show $\rm L^2$RKD is superior to KD, we compare $\rm L^2$RKD with KD and the other state-of-the-art approaches and evaluate its compatibility with the existing approaches.
Extensive experiments on four benchmark datasets with various modern network architectures demonstrate the effectiveness and superiority of $\rm L^2$RKD.


The main contributions of our work are summarized as follows:
\begin{itemize}

\item We observe that only fitting the data points in the vicinity around training data cannot enable a student to well capture the local shape of a teacher function.
On the other hand, we notice that the existing literature ignores the advantage of knowledge distillation that for any given input samples, the teacher function can always output the corresponding soft labels.
In light of these, we propose $\rm L^2$RKD that transfers the knowledge in local, linear regions from a teacher to a student.
To the end, the teacher-student shape difference is significantly reduced and thus the student performance is substantially improved.

\item We compare $\rm L^2$RKD with KD and more than 10 state-of-the-art approaches with various network architectures on four benchmark datasets, i.e., CIFAR-10, CIFAR-100, Tiny ImageNet, and ImageNet.
Despite its simplicity, $\rm L^2$RKD outperforms these approaches substantially and shows large superiority under few-shot settings.
Moreover, $\rm L^2$RKD is compatible with the existing distillation approaches to further improve their performances significantly.

\item Different from the existing work focusing on using different criteria to align logits or representations between teachers and students, we address knowledge distillation from a novel (data) perspective, which provides a promising direction for future work on knowledge distillation.

\end{itemize}

\section{Related Work}
The existing knowledge distillation approaches train students by using both ground truth and supervision from teachers.
Thus, their objectives can be simply expressed as a combination of the regular cross-entropy objective and a distillation objective.
According to the distillation objective, the existing literature can be mainly divided into two categories, i.e., logit-based approaches and representation-based approaches.

\subsection{Logit-based Distillation Approaches}
Logit-based approaches \cite{hinton2015distilling} construct the distillation objective based on output logits.
Hinton et al. \cite{hinton2015distilling} propose KD which uses the softened logits of a teacher network as the targets to train a student network.
The soft targets contain the information about instance-to-class similarities (i..e, dark knowledge) that can improve the student performance.
Park et al. \cite{park2019relational} point out that KD only considers knowledge transfer on individual samples and thus they propose to transfer mutual relations of data examples from a teacher to a student by penalizing logit-based structural differences between them.
Zhao et al. \cite{zhao2020highlight} consider the information in training process for knowledge distillation by employing two teachers.
One teacher uses its temporary output logits during the training process to supervise the student step by step, which assists the student to find the optimal path towards the final logits.
The other teacher provides the information about a critical region that is more useful for the task.
\par

\subsection{Representation-based Distillation Approaches}
It is well observed that DNNs are good at learning different levels of feature representations \cite{butepage2017deep,gidaris2018unsupervised,zhang2018multiview}.
As teacher networks usually have excellent generalization abilities, it is natural to suppose that they have also learned excellent feature representations for the task. 
Based on this idea, representation-based approaches have been proposed.
These approaches design the distillation objective by using the feature representations in intermediate or last several layers.
Fitnets \cite{romero2014fitnets} first uses the intermediate features of a teacher to train a student.
This idea is to directly align the features between the teacher and the student through regressions.
Inspired by this work, Zagoruyko and Komodakis \cite{zagoruyko2016paying} propose to transfer feature based attention maps from a teacher to a student.
Huang and Wang \cite{huang2017like} propose to train a student by aligning the distributions of neuron selectivity patterns between the student
and the teacher.
FSP \cite{yim2017gift} transfers the information in the second order statistics between different layers of a teacher to a student.
Passalis and Tefas \cite{passalis2018learning} propose to transfer knowledge by minimizing the probability distribution differences in the feature space between a student and a teacher.
KDGAN \cite{wang2018kdgan} uses adversarial training strategy to align the representations between a student and a teacher.
AB \cite{heo2019knowledge} transfers the activation boundary from a teacher to a student.
The activation boundary is obtained from pre-activation feature maps.
IRG \cite{liu2019knowledge} considers instance features and instance relationships for knowledge transfer.
CRD \cite{tian2019contrastive} proposes to maximize the mutual information between student and teacher representations.
Other representation-based approaches \cite{pmlr-v80-srinivas18a, heo2019comprehensive, zhu2017feature, cho2019efficacy, ahn2019variational, zhang2019efficient, koratana2019lit, xu2020knowledge, aguilar2019knowledge} utilizes different criteria to align the feature representations between a teacher and a student.
Online knowledge distillation \cite{zhang2018deep, chen2019online, anil2018large, chung2020feature, zhu2018knowledge} has also been proposed to train a group of students simultaneously by aligning their logits or representations.
\par

We observe that the existing studies focus on designing different criteria to align representations or logits between a teacher and a student.
In this work, we address knowledge distillation from a novel data perspective by transferring the knowledge in local, linear regions.


\section{Our Framework}
In this section, we first analyze the convergence of KD from function fitting perspective.
Then, we present the potential issues of KD.
Lastly, we describe $\rm L^2$RKD.

\subsection{KD as Function Fitting}
Given training dataset $\mathcal{T}=(X, Y)$ where $X=\{x_i\}_{i=1}^n$ are training samples drawn from distribution $P(x)$ and $Y=\{y_i\}_{i=1}^n$ are their labels drawn from distribution $P(y|x)$, a teacher function $f_T$ is trained over training data $(X, Y)$ and achieves a good generation error:
\begin{equation}
\label{r}
\mathcal{G}(f_T) =  \mathbb{E}_{P(x,y)}\left[g(f_T(x), y)\right]
\end{equation}
where $g(.)$ measures the error between the prediction and the label.


Hinton et al. \cite{hinton2015distilling} propose KD that minimizes the output probability differences between student $f_S$ and teacher $f_T$ over training dataset $(X, Y)$.
The complete objective is written as:
\begin{equation}
\label{1}
\mathcal{L}_{KD} =   \frac{1}{n}\sum_{i=1}^{n}\lbrack \alpha \mathcal{L}_{CE}(f_S, x_i, y_i) +(1-\alpha) \mathcal{L}_{KL}(f_S, f_T, x_i)\rbrack
\end{equation}
where $\alpha$ is a weight for balancing the contributions of these two terms. $\mathcal{L}_{CE}$ is the regular cross-entropy objective:

\begin{equation}
\label{2}
\mathcal{L}_{CE}(f_S, x, y) = H(y, \sigma(f_S(x))) 
\end{equation}
where $H(.)$ is the cross-entropy loss and $\sigma$ is the softmax function.\par

$\mathcal{L}_{KL}$ in (\ref{1}) is the distillation objective for transferring knowledge from a teacher to a student:
\begin{equation}
\label{3}
\mathcal{L}_{KL}\left(f_S, f_T, x\right) = \tau^2 KL\left(\sigma\left(\frac{f_T\left(x\right)}{\tau}\right), \sigma\left(\frac{f_S\left(x\right)}{\tau}\right)\right) 
\end{equation}
where $\tau$ is a temperature to generate soft labels and $KL$ represents Kullback-Leibler Divergence (KL-divergence) \cite{kullback1951information}.
To investigate the behavior of this distillation objective, we observe its gradient with respect to the student logits $f_S(x)$:
\begin{equation}
\label{e4}
\frac{\partial{\mathcal{L}_{KL}}}{\partial{f_S(x)}} = \tau\left( \sigma\left(\frac{f_S\left(x\right)}{\tau}\right) - \sigma\left(\frac{f_T\left(x\right)}{\tau}\right)\right)
\end{equation}
When $\tau$ is large compared with the magnitude of the logits, the gradient can be approximated (based on Taylor series) as:
\begin{equation}
\label{e5}
\frac{\partial{\mathcal{L}_{KL}}}{\partial{f_S\left(x\right)}} \approx  \tau \left( \frac{1+\frac{f_S\left(x\right)}{\tau}} { K+ \sum_j \frac{f_S\left(x\right)[j]}{\tau} }   - 
\frac{1+\frac{f_T\left(x\right)}{\tau}}{ K+\sum_j \frac{f_T\left(x\right)[j]}{\tau} } \right)
\end{equation}
where $K$ is the total number of classes and $f_S\left(x\right)[j]$ represents the $j$th element of $f_S\left(x\right)$.
With the assumption that the logits of a DNN have mean zero, we can obtain: 
\begin{equation}
\label{fit}
\frac{\partial{\mathcal{L}_{KL}}}{\partial{f_S(x)}} \approx  \frac{1}{K}\left(f_S\left(x\right)-f_T\left(x\right)\right)
\end{equation}
As seen from the gradient, minimizing $\mathcal{L}_{KL}\left(f_S, f_T, x\right)$ with a high temperature is equivalent to minimizing the mean square error between $f_S\left(x\right)$ and $f_T\left(x\right)$.
Thus, the distillation objective in KD can be considered as using one function (student $f_S$) to fit another fixed function (teacher $f_T$).
\par

\subsection{Convergence Analysis of KD}
\label{conver}
As analyzed above, the distillation objective in KD is equivalent to function fitting.
Thus, the goal of distillation is to minimize the difference between student function $f_S$ and teacher function $f_T$ over data distribution $P(x)$:
\begin{equation}
\mathcal{I}_(f_S, f_T, p) = \mathbb{E}_{P(x)}\left[d(f_T(x), f_S(x))\right]
\end{equation}
where $d(.)$ measures the difference between two vectors.
If $\mathcal{I}_(f_S, f_T, p)$ is small enough, then the student $f_S$ is almost as accurate as the teacher $f_T$, i.e., their generalization errors are very close: $\mathcal{G}(f_S)\approx \mathcal{G}(f_T)$.

However, data distribution $P(x)$ is typically unknown in practice.
Instead, KD minimizes the empirical difference:
\begin{equation}
\label{difem}
\mathcal{I}_{emp}(f_S, f_T, X) = \frac{1}{n}\sum_{i=1}^{n}\left[d(f_T(x_i), f_S(x_i))\right]
\end{equation}
With uniform convergence bounds \cite{vapnik1998statistical}, minimizing $\mathcal{I}_{emp}$ for $f_S \in S$ leads to an fitting error that can be bounded as $O(\sqrt{\frac{C}{n}})$ where C denotes the complexity of function class $S$ and $n$ is the number of samples used for distillation.
This bound suggests that using more \textbf{unlabeled data} for distillation can reduce this fitting error.
This is ignored by the existing studies as they focus on using different criteria to align the logits or representations between teacher and student networks.
In this paper, we make use of this hint and address knowledge distillation from a novel data perspective.
The challenge is that we do not have the real data distribution $P(x)$ in practice to draw additional, unlabeled samples.

\subsection{Locally Linear Region Knowledge Distillation}
Training $f_S$ by minimizing (\ref{1}) is known as the {\em Empirical Risk Minimization} (ERM) principle \cite{book}.
While efficient to compute, the empirical risk (\ref{1}) only guides student $f_S$ to fit teacher $f_T$ at a finite set of sparse training data points.
As analyzed above, when we only use sparse training data points for distillation (i.e., $n$ is small), the fitting error is large.
In this case, the student cannot sufficiently capture the knowledge in the teacher, which results in a large teacher-student performance gap.
To alleviate this issue, data augmentation based on the VRM principle is typically used.
In VRM, new samples are drawn from a human defined vicinity distribution of the training data:
\begin{equation}
\label{9}
P_v(x^{aug},y^{aug})=\frac{1}{n}\sum^n_{i=1} v(x^{aug},y^{aug}\big | x_i,y_i)
\end{equation}
where $v$ is a vicinity distribution that measures the probability of $(x^{aug},y^{aug})$ existing in the neighborhood of training data point $(x_i,y_i)$.
In practice, we usually only augment each training sample $x_i$ without changing label $y_i$.
For instance, in image classification, $x^{aug}_i$ can be obtained by slightly padding and randomly cropping $x_i$ and remaining label $y_i$.\par

\begin{figure}[!t]
\centering
     \includegraphics[height=0.23\textwidth]{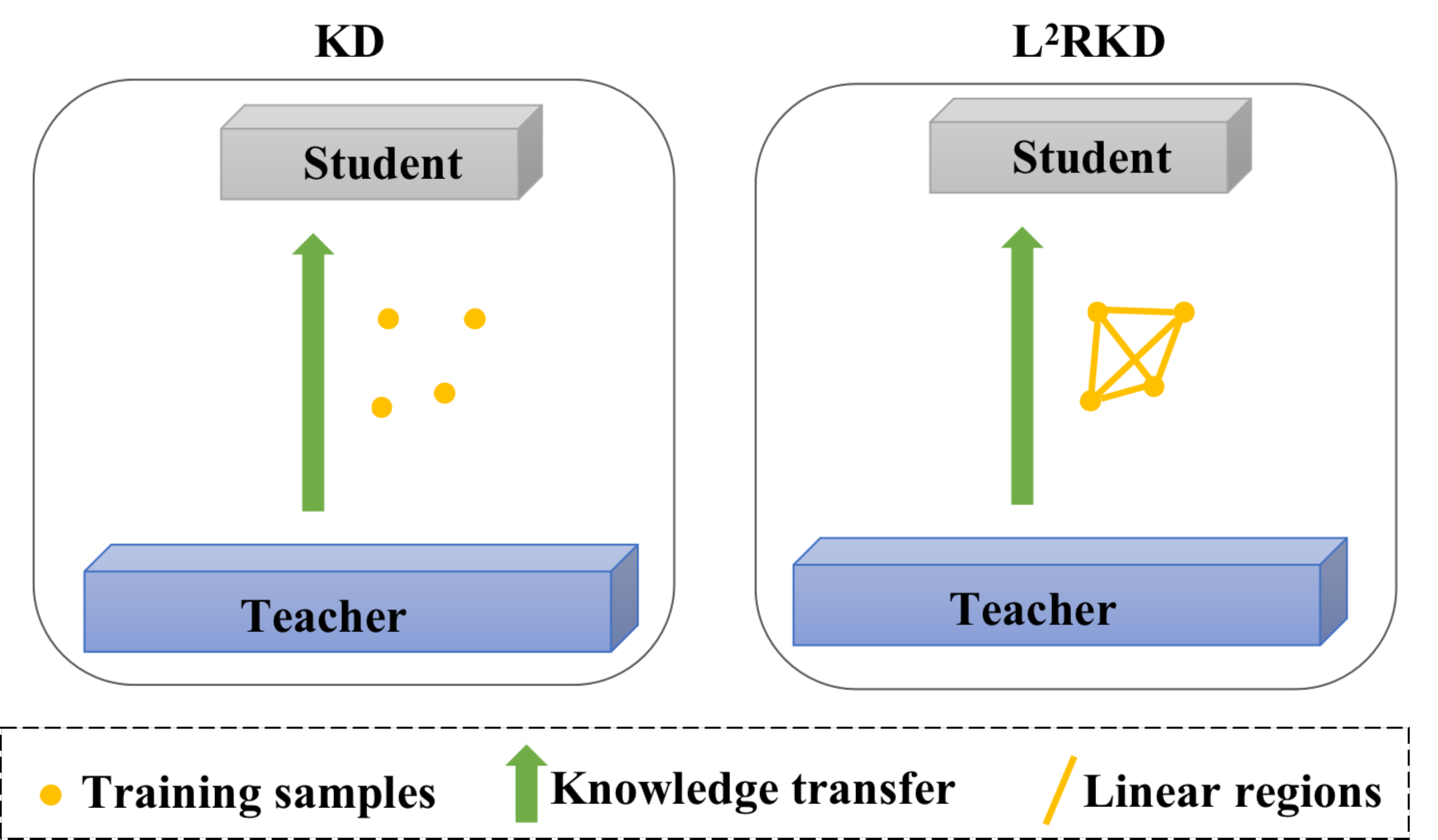}
     \caption{Differences between $\rm L^2$RKD and KD}
     \label{frame}
\end{figure}

By drawing samples from $P_v(x^{aug},y^{aug})$, a new training dataset is obtained, i.e., $\mathcal{T}_v$=\{$(x^{aug1}_1,y_1)$,$(x^{aug2}_1,y_1)$, ..., $(x^{augm}_1,y_1)$, $(x^{aug1}_2,y_2)$,$(x^{aug2}_2,y_2)$, ..., $(x^{augm}_2,y_2)$, ..., $(x^{aug1}_n,y_n)$, $(x^{aug2}_n,y_n)$, ..., $(x^{augm}_n,y_n)$\} where $(x^{aug1}_i,y_i)$, $(x^{aug2}_i,y_i)$, ..., $(x^{augm}_i,y_i)$ are $m$ samples drawn from the vicinity of $(x_i, y_i)$ ($m$ can be considered as the total training epochs in SGD optimization).
This leads to the {\em empirical vicinal risk} of KD:

\begin{equation}
\label{10}
\begin{aligned}
\mathcal{L}_{v} =\frac{1}{mn}\sum_{i=1}^{n}
\sum_{j=1}^{m}[\alpha \mathcal{L}_{CE}(x_i^{augj}, y_i) \\+(1-\alpha) \mathcal{L}_{KL}(x_i^{augj})]
\end{aligned}
\end{equation}
This objective is the most commonly used KD objective in practice.
However, the limitation is that the new drawn data samples from the vicinity distribution is very close to the original training data points, since they contain almost the identical objective only with different backgrounds caused by padding or cropping.
As shown in Figure \ref{f2}, with data augmentation, the student can better capture the local shape of the teacher, but still only covers a very small region around each training sample $x_i$.


To address this issue, we propose $\rm L^2$RKD that enforces the student to mimic the behavior of the teacher in local, linear regions. 
The differences between $\rm L^2$RKD and KD are shown in Figure \ref{frame}.
Instead of transferring knowledge in sparse training data points, $\rm L^2$RKD
transfers the knowledge in local, linear regions from the teacher to the student.
We denote the distribution of the points in the linear regions between two training or augmented samples by $P_{lr}(\hat x)$.
Then the objective of $\rm L^2$RKD can be expressed as:
\begin{equation}
\label{lr}
\begin{aligned}
\mathcal{L}_{lr} =\frac{1}{mn}\sum_{i=1}^{n}
\sum_{j=1}^{m}[\alpha \mathcal{L}_{CE}(f_S, x_i^{augj}, y_i)]+\\ \eta*\mathbb{E}_{\hat x\sim P_{lr}(\hat x)}\left[\tau^2 KL\left(\sigma\left(\frac{f_T\left(\hat x\right)}{\tau}\right), \sigma\left(\frac{f_S\left(\hat x\right)}{\tau}\right)\right)\right]
\end{aligned}
\end{equation}
where $\alpha$ and $\eta$ are two weights for balancing the contributions of the regular cross-entropy loss and the distillation loss.
\par

\begin{figure}[!t]
\centering
     \includegraphics[height=0.23\textwidth]{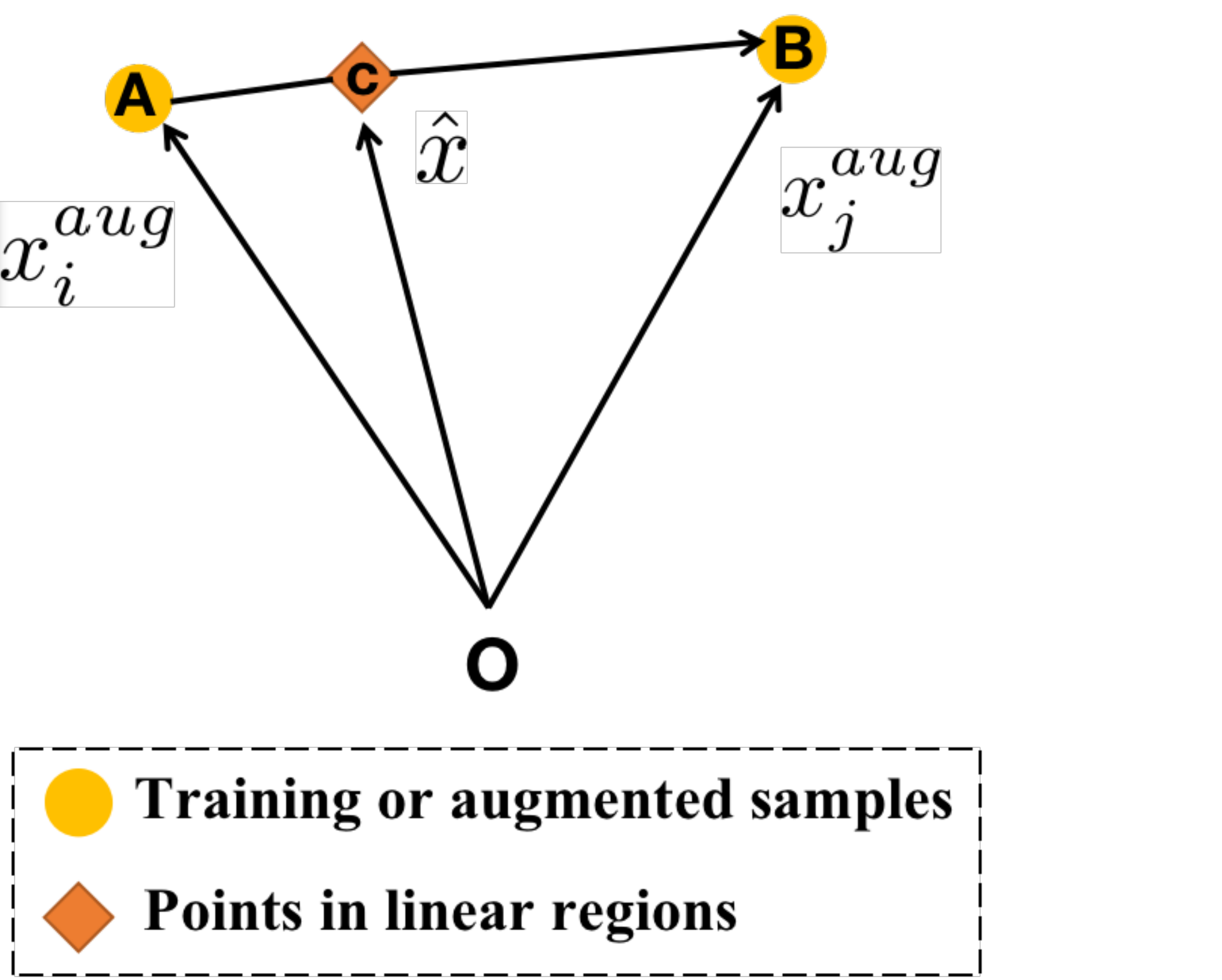}
     \caption{Points in linear regions}
     \label{vector}
\end{figure}

To optimize (\ref{lr}), we draw points from linear regions (i.e., $P_{lr}$) by linear algebra.
As shown in Figure \ref{vector}, each data sample (e.g., $x_i^{aug}$ and $x_j^{aug}$) can be considered as a high-dimensional vector:
\begin{equation}
x_i^{aug} =\overrightarrow{OA},\quad x_j^{aug} =\overrightarrow{OB}
\end{equation}
$\hat x$ is a random point in the linear region between two augmented samples, i.e., $x_i^{aug}$ and $x_j^{aug}$.
We can obtain $\hat x$ by using vector operations:
\begin{equation}
\overrightarrow{AB} =\overrightarrow{OB}-\overrightarrow{OA};\quad
\overrightarrow{OC} =\overrightarrow{OA}+\lambda*\overrightarrow{AB};\quad \hat x=\overrightarrow{OC}
\end{equation}
where $\lambda \in [0, 1]$ controls which point we sample from the linear region.
Note that local, linear regions contain both in-distribution points and out-of-distribution points, which indicates that $\rm L^2$RKD goes beyond in-distribution distillation.
This is also the main difference between $\rm L^2$RKD and regular data augmentation as regular data augmentation only generates in-distribution samples with ground truth.

\par

The implementation-level description of $\rm L^2$RKD is presented in Algorithm \ref{g1}.

    \begin{algorithm}[!t]  
        \caption{$\rm L^2$RKD}  
        \label{g1}
        \begin{algorithmic}[1] 
         \renewcommand{\algorithmicrequire}{\textbf{Input:}} 
          \renewcommand{\algorithmicensure}{\textbf{Output:}}
            \Require Training data $(X, Y)$, teacher $f_T$, and student $f_S$
            \Ensure Student $f_S$ with optimal parameters 
            \For{$i = 1, 2, ..., k $}  
                    \State Randomly take two mini-batch of training data ($x_{batch}^{i}, y_{batch}^{i}$) and ($x_{batch}^{j}, y_{batch}^{j}$) from $(X, Y)$
                    \State Generate $x_{batch}^{augi}$ and $x_{batch}^{augj}$ by using data augmentation (e.g., randomly flipping, padding, and cropping)
                    \State Randomly sample $\lambda$ from [0, 1]
                    \State Obtain 
                    $\hat x_{batch}$ by (14) by using $x_{batch}^{augi}$ and $x_{batch}^{augj}$
                    \State With $\hat x_{batch}$, minimize (12) with one gradient descent step for $f_S$
                \EndFor  

        \end{algorithmic}  
    \end{algorithm}
\subsection{Analysis of $L^2$RKD}
$\rm L^2$RKD distills the knowledge in local, linear regions instead of sparse data points.
This is achieved by enforcing students to mimic the teacher shapes in local, linear regions as shown in Figure \ref{f3}.
One may consider $\rm L^2$RKD as a data augmentation strategy, but there are many worth-noting differences.
Regular data augmentation usually generates in-distribution samples with ground truth, i.e., $(x, y)$ where $(x, y)$ follows the real data distribution $P(x, y)$.
These generated pairs are used to learn a function $f$ mapping $x$ to $y$ where $(x, y)\sim P(x, y)$.
In contrast, $\rm L^2$RKD aims to enforce a student $f_S$ to fit the shape of a teacher function $f_T$ in local, linear regions.
These local, linear regions contain in-distribution samples \textbf{without ground truth} and out-of-distribution samples.
These samples assist a student $f_S$ to learn the distribution $Q(x, z)$ defined by teacher $f_T$ where $x$ are the inputs and $z$ are the output probabilities of $f_T$.
Note that $Q(x, z)$ is different from $P(x, y)$ as $f_t$ is usually not perfect.
\textbf{The advantage of fitting $Q(x, z)$ is that \textbf{$Q(x, z)$ is more tractable than $P(x, y)$}}, since for given any sample $x$, $f_T$ can always give output $f_T(x)$ and $(x, \sigma(\frac{f_T(x)}{\tau}))$ follows $Q(x, z)$.
Even for out-of-distribution samples in the linear regions, $f_T$ can still output soft labels.
It is obvious that these soft labels for out-of-distribution samples are semantically meaningless.
However, by querying $f_T$ in local, linear regions, the student is able to better explore the local shape of the teacher, thus obtaining more performance gain.
This advantage is ignored by the existing work.
In contrast, $\rm L^2$RKD makes use of this advantage by using freely obtained local, linear regions to better capture the knowledge in the teacher.


\section{Experiments}

We report extensive experiments for evaluating $\rm L^2$RKD.
We first conduct ablation study regarding local linear regions.
Then we show that $\rm L^2$RKD is superior to KD by (1) comparing $\rm L^2$RKD with KD and the other state-of-the-art approaches, (2) demonstrating that $\rm L^2$RKD is more compatible with the existing approaches, (3) demonstrating the superiority of $\rm L^2$RKD under few-shot settings, (4) empirically verifying that the local shapes of the students train with $\rm L^2$RKD are more closer to the teachers' shapes than those of the students trained with KD.
Last but not least, we compare $\rm L^2$RKD with other data-driven techniques.


\begin{figure}[!t]
\centering
     \includegraphics[height=0.35\textwidth]{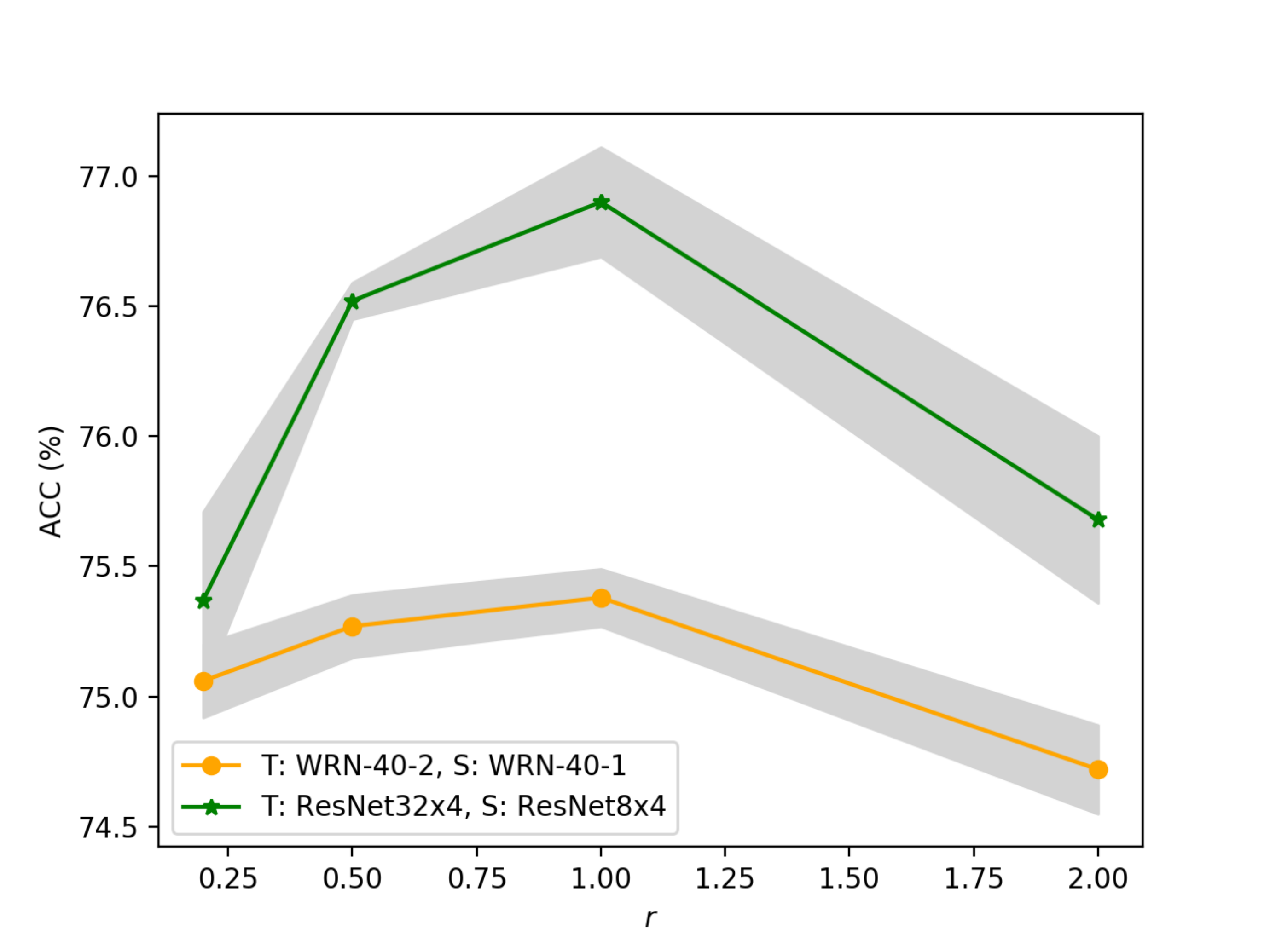}
     \caption{Ablation studies regarding $r$ on CIFAR-100}
     \label{ab}
\end{figure}

\begin{table*}[!t]
\setlength{\abovecaptionskip}{0.1cm}
\setlength{\belowcaptionskip}{0.1cm}
\caption{Test accuracies (\%) on CIFAR-10}
\centering
\label{m1}
\begin{tabular}{lccccccc}
\toprule
Teacher    & WRN-40-2           & ResNet32$\times$4    & WRN-16-4   & ResNet-110 &ResNet-56   & VGG-13           & ResNet-110     \\
Student    & WRN-16-1            & ResNet8$\times$4    & WRN-16-1  & MobileNetV2  & WRN-16-1   & MobileNetV2      & WRN-16-1       \\   \midrule
Teacher    & 94.80                  & 95.52&      95.04& 94.00       & 93.87      & 94.01            & 94.00                 \\
Vanilla Student    & 91.32              & 92.78&  91.32&   89.19    &  91.32      & 89.19              & 91.32            \\ \midrule
KD        & 91.77$\pm$0.25     & 93.85$\pm$0.22& 92.62$\pm$0.20& 89.61$\pm$0.26 & 91.86$\pm$0.18  & 89.21$\pm$0.17     & 91.52$\pm$0.16       \\
$\rm L^2$RKD & \textbf{93.11$\pm$0.21} & \textbf{94.75$\pm$0.19}&\textbf{93.24$\pm$0.25}& \textbf{90.83$\pm$0.18}   & \textbf{92.64$\pm$0.21} & \textbf{90.52$\pm$0.11} & \textbf{ 92.47$\pm$0.07 } \\ \midrule
FitNet   & 91.20$\pm$0.12     & 93.21$\pm$0.29& 90.77$\pm$0.34 &88.84$\pm$0.25     & 89.71$\pm$0.41  & 87.51$\pm$0.21  & 91.15$\pm$0.15         \\
AT         & 91.96$\pm$0.29     & 93.45$\pm$0.19& 91.88$\pm$0.10&89.40$\pm$0.33    & 91.24$\pm$0.23   & 87.06$\pm$0.09          & 91.41$\pm$0.26  \\
SP         & 91.78$\pm$0.18     & 93.31$\pm$0.15& 92.19$\pm$0.31&89.09$\pm$0.36   &91.27$\pm$0.23    & 88.55$\pm$0.28          & 91.07$\pm$0.27      \\
CC         & 91.00$\pm$0.26      &92.95$\pm$0.32&  91.01$\pm$0.35&88.13$\pm$0.41 & 90.96$\pm$0.12     & 87.34$\pm$0.15       & 91.04$\pm$0.10    \\
VID        & 91.47$\pm$0.20     & 93.65$\pm$0.25&  90.99$\pm$0.11&89.50$\pm$0.22 &90.94$\pm$0.39        & 87.23$\pm$0.25    & 91.56$\pm$0.28          \\
RKD        & 91.50$\pm$0.25     & 93.20$\pm$0.19& 92.25$\pm$0.31& 89.20$\pm$0.17 &91.48$\pm$0.11      & 88.45$\pm$0.27  & 91.32$\pm$0.12         \\
AB        & 91.01$\pm$0.18   & 92.88$\pm$0.28&    90.70$\pm$0.51& 88.37$\pm$0.19 &90.61$\pm$0.40    & 87.81$\pm$0.29       & 90.52$\pm$0.26         \\
PKT        & 91.97$\pm$0.17     & 94.09$\pm$0.27& 92.28$\pm$0.08&89.18$\pm$0.15  &91.64$\pm$0.26      & 88.60$\pm$0.12    & 91.91$\pm$0.14        \\
CRD        & 91.95$\pm$0.28     & 94.08$\pm$0.17& 89.67$\pm$0.48 &85.70$\pm$0.10 & 90.37$\pm$0.22       & 89.16$\pm$0.12     & 91.80$\pm$0.21         \\
FT      & 91.01$\pm$0.25     & 92.97$\pm$0.15&  91.30$\pm$0.09& 80.56$\pm$0.41  &90.69$\pm$0.21     & 83.44$\pm$0.19       & 90.92$\pm$0.05        \\
NST       & 91.56$\pm$0.11     & 92.92$\pm$0.20&  91.14$\pm$0.24&88.77$\pm$0.34  & 90.78$\pm$0.12   & 82.95$\pm$0.42        &  90.66$\pm$0.15     \\ \bottomrule
\end{tabular}
\end{table*}

\subsection{Datasets}
In the experiments, we adopt four benchmark datasets: CIFAR-10 \cite{krizhevsky2009learning}, CIFAR-100 
\cite{krizhevsky2009learning}, Tiny ImageNet \footnote{http://tiny-imagenet.herokuapp.com/}, and ImageNet (ILSVRC2012) \cite{deng2009imagenet}.

$\textbf{CIFAR-10}$ is an image classification dataset with 10 classes, containing 50,000 training images and 10,000 test images with image size 32 $\times$ 32 in the RGB space.
We follow the standard data augmentation on CIFAR datasets.
During training time, we pad 4 pixels on each side of an image and randomly flip it horizontally.
Then the image is randomly cropped to 32 $\times$ 32 size. 
During test time, we only evaluate the single view of an original 32 $\times$ 32 image without padding or cropping.

$\textbf{CIFAR-100}$ comprises similar images to those in CIFAR-10, but has 100 classes.
We adopt the same data augmentation strategy as that in CIFAR-10.

$\textbf{Tiny ImageNet}$, i.e., a subset of ImageNet, is an image classification dataset with 200 classes, containing 100,000 training images and 10,000 test images with size 64 $\times$ 64 in the RGB space.
At training time, we pad 8 pixels on each side of an image and randomly flip it horizontally, then the image is randomly cropped to 64 $\times$ 64 size.
At test time, we only evaluate the original image.

$\textbf{ImageNet}$ is a large-scale image classification dataset with 1000 classes, containing 1.28 million training images and 50,000 validation images with different sizes in the RGB space.
On ImageNet, we use the standard scale and aspect ratio augmentation strategy from \cite{szegedy2015going}.
Test images are resized so that the shorter side is set to 256, and then are cropped to size 224 $\times$ 224.\par

\subsection{Network Architectures and Hypermeters}
To evaluate the applicability of $\rm L^2$RKD on different network architectures, we adopt various modern architectures, i.e., ResNet \cite{he2016deep}, WRN \cite{Zagoruyko2016WRN}, VGG \cite{Simonyan15}, MobileNet \cite{sandler2018mobilenetv2}, and ShuffleNet \cite{ma2018shufflenet}.
\par

On CIFAR and  Tiny ImageNet datasets, $\alpha$, $\eta$, and $\tau$ are set to 0.1, 1, and 4, respectively.
On ImageNet, $\alpha$, $\eta$, and $\tau$ are set to 1, 1, and 2, respectively.
All the networks are trained with SGD with momentum 0.9 for 240, 100, and 120 epochs on CIFAR, Tiny ImageNet, and ImageNet, respecitively.
The min-batch size is set to 64, 64, and 256 for CIFAR, Tiny ImageNet, and ImageNet, respecitively.
For ResNet, WRN, and VGG, we set the initial learning rate to 0.05 on CIFAR and Tiny ImageNet.
For MobileNet and ShuffleNet, the initial learning rate is set to 0.01 on CIFAR and Tiny ImageNet.
On CIFAR, the learning rate is divided by 10 every 30 (40 for WRN-16) epochs after the first 150 (120 for WRN-16) epochs.
For Tiny ImageNet, the learning rate is divided by 5 every 30 epochs.
On ImageNet, the initial learning rate is set to 0.1 and is divided by 10 every 30 epochs.

\par

\subsection{Competitors}
We compare $\rm L^2$RKD with KD \cite{hinton2015distilling} and several other state-of-the-art approaches, i.e., FitNet \cite{romero2014fitnets}, AT \cite{zagoruyko2016paying}, SP \cite{tung2019similarity}, FT \cite{kim2018paraphrasing}, NST \cite{huang2017like}, CC \cite{peng2019correlation}, FSP \cite{yim2017gift}, PKT \cite{passalis2018learning}, AB \cite{heo2019knowledge}, VID \cite{ahn2019variational}, RKD \cite{park2019relational}, and CRD \cite{tian2019contrastive}.
For these approaches, their objectives can be expressed as a combination of the regular cross-entropy loss and a distillation loss:
\begin{equation}
\label{8}
\mathcal{L} = \mathcal{L}_{CE} + c  \mathcal{L}_{distill}
\end{equation}
where $c$ is a weight for balancing the two terms.
We report the author-reported results, or use author-provided codes and the optimal hyper-parameters from the original papers if they are publicly available.
Otherwise, we use the public implementation of \cite{tian2019contrastive}.
Specifically, the optimal hyper-parameters for each method are: (1) FitNet: $c=100$; (2) AT: $c=1000$; (3) SP: $c=3000$; (4) CC: $c=0.02$; (5) VID: $c=1$; (6) RKD: $c_1 = 25$ for the distance metric and $c_2 = 50$ for the angle metric; both terms are combined following the original paper; (7) PKT: $c=30000$; (8) AB: $c = 0$; distillation happens in the pre-training stage where only the distillation objective is used; (9) FT: $c=500$; (10) CRD: $c=0.8$; (11) FSP: $c=0$; distillation happens in the pre-training stage where only the distillation objective is used; (12) NST: $c=50$; (13) the KD objective is (\ref{10}); $\alpha$ and $\tau$ are set to 0.1 and 4, respectively.\par
All the results below are reported based on 3 runs.

\subsection{Ablation Studies regarding Local, Linear Regions}
The local, linear regions contain much more data points than those in the training dataset.
In the training process, $\rm L^2$RKD randomly draws samples from these local regions for knowledge transfer.
Thus, we first conduct an ablation study regarding the number of the drawn points used for distillation.
We use $r$ to denote the ratio of the number of training samples to the number of samples drawn from the local regions.
Specifically, we adopt two pairs of teacher and student networks on CIFAR-100, i.e., teacher WRN-40-2 and student WRN-40-1, and teacher ResNet32$\times$4 and student ResNet8$\times$4.
We choose the ratio $r\in\{0.2, 0.5, 1, 2\}$.\par

Figure \ref{ab} presents the performances of $\rm L^2$RKD with different $r$ values on CIFAR-100.
We observed that the performance of $\rm L^2$RKD is positively related to the value of $r$ when $r$ is less than a threshold.
It means that drawing more samples from the local, linear regions is beneficial to the student performance when the number of the drawn samples is less than a threshold.
However, further increasing the number of the drawn points hurts the performance.
The best performance for both teacher-student pairs is obtained at $r=1$.
Thus, we simply set $r$ to 1 in the following experiments.
$r$ = 1 means that we sample the same number of points from the local, linear regions as that of the training data points.

\begin{table*}[!t]
\setlength{\abovecaptionskip}{0.1cm}
\setlength{\belowcaptionskip}{0.1cm}
\caption{Test accuracies (\%) on CIFAR-100}
\centering
\label{m2}
\begin{tabular}{lccccccccc}
\toprule
Teacher    & VGG-13  & ResNet32$\times$4 & WRN-40-2 & WRN-40-2 & ResNet-56   & ResNet-110 &WRN-40-2 &ResNet32$\times$4 \\
Student    & VGG-8 & ResNet8$\times$4  & WRN-16-2   & WRN-40-1    & ResNet-20    & ResNet-32  &VGG-8 &ShuffleNetV2        \\ \midrule
Teacher    & 74.64  & 79.42             & 75.61               & 75.61               & 72.34              & 74.31  & 75.61    & 79.42             \\
Vanilla Student    & 70.36               & 72.50               & 73.26               & 71.98               & 69.06               & 71.14  &  70.36  &  71.86       \\ \midrule
KD         & 72.98$\pm$0.19          & 73.33$\pm$0.25          & 74.92$\pm$0.28          & 73.54$\pm$0.20          & 70.66$\pm$0.24            & 73.08$\pm$0.18  & 73.51$\pm$0.17  & 74.45$\pm$0.27        \\
$\rm L^2$RKD & \textbf{74.54$\pm$0.15} & \textbf{75.67$\pm$0.18} & \textbf{75.91$\pm$0.22} & \textbf{75.38$\pm$0.11} &  \textbf{71.30$\pm$0.19} & \textbf{73.61$\pm$0.21} & \textbf{75.05$\pm$0.13} &  \textbf{76.90$\pm$0.21} \\ \midrule
FitNet     & 71.02$\pm$0.31         & 73.50$\pm$0.28          & 73.58$\pm$0.32          & 72.24$\pm$0.24          & 69.21$\pm$0.36                & 71.06$\pm$0.13     &71.14$\pm$0.17. & 73.54$\pm$0.22     \\
AT         & 71.43$\pm$0.09          & 73.44$\pm$0.19          & 74.08$\pm$0.25          & 72.77$\pm$0.10          & 70.55$\pm$0.27               & 72.31$\pm$0.08  &70.30$\pm$0.21  & 72.73$\pm$0.09        \\
SP         & 72.68$\pm$0.19          & 72.94$\pm$0.23          & 73.83$\pm$0.12          & 72.43$\pm$0.27          & 69.67$\pm$0.20               & 72.69$\pm$0.41    &73.12$\pm$0.18  & 74.56$\pm$0.22      \\
CC         & 70.71$\pm$0.24          & 72.97$\pm$0.17          & 73.56$\pm$0.26          & 72.21$\pm$0.25          & 69.63$\pm$0.32               & 71.48$\pm$0.21    &70.64$\pm$0.20  &71.29$\pm$0.38      \\
VID        & 71.23$\pm$0.06          & 73.09$\pm$0.21          & 74.11$\pm$0.24          & 73.30$\pm$0.13          & 70.38$\pm$0.14               & 72.61$\pm$0.28      &71.86$\pm$0.23 & 73.40$\pm$0.17     \\
RKD        & 71.48$\pm$0.05      & 71.90$\pm$0.11          & 73.35$\pm$0.09          & 72.22$\pm$0.20          & 69.61$\pm$0.06                 & 71.82$\pm$0.34      &71.00$\pm$0.19   & 73.21$\pm$0.28    \\
PKT        & 72.88$\pm$0.09          & 73.64$\pm$0.18          & 74.54$\pm$0.04          & 73.45$\pm$0.19          & 70.34$\pm$0.04                & 72.61$\pm$0.17      &72.74$\pm$0.42 &74.69$\pm$0.34    \\
AB         & 70.94$\pm$0.18          & 73.17$\pm$0.31          & 72.50$\pm$0.26          & 72.38$\pm$0.31          & 69.47$\pm$0.09                 & 70.98$\pm$0.39      & 72.21$\pm$0.41  & 74.31$\pm$0.11     \\
FT         & 70.58$\pm$0.08         & 72.86$\pm$0.12          & 73.25$\pm$0.20          & 71.59$\pm$0.15          & 69.84$\pm$0.12                & 72.37$\pm$0.31      &68.33$\pm$0.22    & 72.50$\pm$0.15    \\
CRD & 73.94$\pm$0.22          &75.51$\pm$0.18    &  75.48$\pm$0.09  & 74.14$\pm$0.22  & 71.16$\pm$0.17   & 73.48$\pm$0.13   & 74.08$\pm$0.20 & 75.65$\pm$0.10 \\
NST        & 71.53$\pm$0.13          & 73.30$\pm$0.25          & 73.68$\pm$0.11          & 72.24$\pm$0.22          & 69.60$\pm$0.13                & 71.96$\pm$0.07       &69.56$\pm$0.24 & 74.68$\pm$0.26   \\ \bottomrule
\end{tabular}%
\end{table*}

\begin{table*}[!tp]
\caption{Test accuracies (\%) on Tiny ImageNet}
\label{m2_2}
\centering
\begin{tabular}{cc|cc|ccccccccccc}
\toprule
   Teacher   &Student     & KD     & $\rm L^2$RKD &   FitNet & AT      & CC     &SP            & VID            & RKD  & CRD  & PKT    & AB  \\ \midrule
   
    \begin{tabular}{@{}c@{}}VGG-13 \\(61.62)\end{tabular}    &\begin{tabular}{@{}c@{}}VGG-8 \\(55.46)\end{tabular}     & \begin{tabular}{@{}c@{}}60.21 \\$\pm$0.19\end{tabular}  & \begin{tabular}{@{}c@{}} \textbf{62.31} \\\textbf{$\pm$0.16}\end{tabular}     &\begin{tabular}{@{}c@{}}55.26 \\$\pm$0.20\end{tabular} &\begin{tabular}{@{}c@{}}56.82 \\$\pm$0.46\end{tabular}&  \begin{tabular}{@{}c@{}} 54.14 \\$\pm$0.19\end{tabular}   &  \begin{tabular}{@{}c@{}} 56.99 \\$\pm$0.42\end{tabular}     &  \begin{tabular}{@{}c@{}} 54.57 \\$\pm$0.26\end{tabular}     &  \begin{tabular}{@{}c@{}} 56.60 \\$\pm$0.13\end{tabular} &\begin{tabular}{@{}c@{}} 59.95 \\$\pm$0.23\end{tabular} &\begin{tabular}{@{}c@{}} 56.36 \\$\pm$0.17\end{tabular}  &   \begin{tabular}{@{}c@{}}55.41\\$\pm$0.36\end{tabular}        \\  \midrule
  \begin{tabular}{@{}c@{}}WRN-40-2 \\(61.84)\end{tabular}   &\begin{tabular}{@{}c@{}}WRN-40-1 \\(55.39)\end{tabular}   &\begin{tabular}{@{}c@{}}56.25 \\$\pm$0.15\end{tabular}     &\begin{tabular}{@{}c@{}}\textbf{57.60} \\\textbf{$\pm$0.19}\end{tabular}     &\begin{tabular}{@{}c@{}}55.41 \\$\pm$0.31\end{tabular}         &\begin{tabular}{@{}c@{}}55.84\\$\pm$0.41\end{tabular}       &\begin{tabular}{@{}c@{}}55.10 \\$\pm$0.43\end{tabular}      &\begin{tabular}{@{}c@{}} 54.09\\$\pm$0.26 \end{tabular}  &\begin{tabular}{@{}c@{}}56.07 \\$\pm$0.23\end{tabular}    &\begin{tabular}{@{}c@{}}55.37\\$\pm$0.29\end{tabular} & \begin{tabular}{@{}c@{}}56.75\\$\pm$0.33 \end{tabular} & \begin{tabular}{@{}c@{}}56.31\\$\pm$0.22\end{tabular} &\begin{tabular}{@{}c@{}}55.76\\$\pm$0.26\end{tabular}            \\ \bottomrule
  
\end{tabular}%
\end{table*}

\begin{table}[!tp]
\setlength{\belowcaptionskip}{0.1cm}
\setlength{\abovecaptionskip}{0.1cm}
\setlength{\belowcaptionskip}{0.1cm}
\caption{Comparison results on ImageNet}
\centering
\label{m_img}
\resizebox{8.4cm}{!}{
\begin{tabular}{l|ll|ll|llll}
\toprule
      & Teacher                   & Student & KD    & $\rm L^2$RKD   & AT    & CRD   & SP    & CC   \\ \midrule
TOP-1 & \multicolumn{1}{l}{73.31} & 69.75   & 70.66 & \textbf{71.53} & 70.70 & 71.17 & 70.22 & 69.96  \\
TOP-5 & 91.42                      & 89.07  & 89.88  & \textbf{90.55}  & 90.00 & 90.13 & 89.80 & 89.17  \\ \bottomrule
\end{tabular}
}
\end{table}

\subsection{Comparison with State-of-the-art Approaches}
We report the comparison results on four benchmark datasets with different network architectures.
\subsubsection{Comparison Results on CIFAR-10}
Table \ref{m1} summarizes the comparison results on CIFAR-10.
We have the following observations.
First, through distilling the knowledge in local regions, $\rm L^2$RKD outperforms KD significantly and also beats the other state-of-the-art approaches substantially across different network architectures, which demonstrates the effectiveness of $\rm L^2$RKD and indicates the importance of the knowledge in local, linear regions.
Second, the approaches based on logits or the last several layers' features (e.g., KD, PKT, CRD, and $\rm L^2$RKD) perform better than the approaches based on intermediate features (e.g., FitNet and AT).
The reason can be that the output logits and the last several layers' features of the teacher contain more high-level information compared to the features in the shallow layers.
Third, when the teacher architecture and the student architecture are extremely different (e.g., VGG-13 and MobileNetV2), almost all the representation-based approaches (e.g., FitNet and AT) fail to transfer knowledge from the teacher to the student, even underperforming the vanilla student.
This indicates that aligning the features of a teacher and a student with extremely different architectures hurts the student performance.
In contrast, $\rm L^2$RKD can still largely improve the student performance on these architecture-different pairs, which demonstrates the robustness of $\rm L^2$RKD.

\subsubsection{Comparison Results on CIFAR-100}
We further evaluate $\rm L^2$RKD on CIFAR-100.
Table \ref{m2} reports the comparison results on CIFAR-100.
It is observed that $\rm L^2$RKD beats KD and the other approaches significantly.
We also notice that on some teacher-student pairs such as teacher VGG-13 and student VGG-8, teacher WRN-40-2 and student WRN-16-2, and teacher WRN-40-2 and student WRN-40-1, $\rm L^2$RKD even performs on a par with the teachers while the other approaches have an obvious performance gap to the teachers.
This demonstrates the superiority and effectiveness of $\rm L^2$RKD.
On the other hand, we also notice that most of the existing approaches fail to improve the student performance on the pair of teacher WRN-40-2 and student VGG-8.
The reason is that this pair of teacher and student has extremely different architectures, which hinders knowledge transfer.
However, $\rm L^2$RKD still improves the student performance substantially from 72.50 to 75.05, which demonstrates the robustness of $\rm L^2$RKD on teacher-student pairs on which the knowledge is difficult to transfer.

\subsubsection{Comparison Results on Tiny ImageNet}
We further report the the comparison results Tiny ImageNet.
As shown in Table \ref{m2_2}, $\rm L^2$RKD shows consistent improvements over all the other approaches and even makes student VGG-8 (62.31\%) significantly outperform teacher VGG-13 (61.62\%), which demonstrates the superiority of $\rm L^2$RKD.\par


\subsubsection{Comparison Results on ImageNet}
We further investigate the performance of $\rm L^2$RKD  on large scale dataset ImageNet.
Limited by computation resources, we only adopt one teacher-student pair on ImageNet.
We follow \cite{tian2019contrastive} and use ResNet-34 and ResNet-18 as the teacher and the student, respectively.
The comparison results are reported in Table \ref{m_img}.
It is observed that $\rm L^2$RKD improves both top-1 and top-5 accuracies over KD and the other state-of-the-art approaches significantly, which demonstrates the applicability and usefulness of $\rm L^2$RKD on large scale datasets.

\subsection{Compatibility with State-of-the-art Approaches}
Representation-based approaches can be combined with logit-based approaches to further improve their performances.
Currently, state-of-the-art approaches are combined with KD to obtain further performance gain.
In this part, we show that the state-of-the-art approaches combined with $\rm L^2$RKD are able to obtain more performance gain.
\par
The compatibility performances are reported in Table \ref{mcom}.
We observe that the existing approaches combined with $\rm L^2$RKD consistently outperform themselves combined with KD by a large margin on all the settings where the teachers and students share similar architectures (e.g., Resnet32$\times$4 and student ResNet$\times$8) or use different architectures (e.g., ResNet-50 and VGG-8).
For example, with teacher Resnet32$\times$4 and student ResNet8$\times$4, FitNet+$\rm L^2$RKD achieves the test accuracy of 76.36\% that is much higher than that of FitNet+KD i.e., 74.66\%.
This demonstrates that $\rm L^2$RKD is more compatible with the existing approaches than KD and is able to further improve their performances substantially.

\begin{table*}[!t]
\setlength{\abovecaptionskip}{0.1cm}
\setlength{\belowcaptionskip}{0.1cm}
\caption{Compatibility performances on CIFAR-100}
\label{mcom}
\centering
\begin{tabular}{lcccccc}
\toprule
Teacher     & ResNet32$\times$4    & VGG-13  & ResNet-50    & ResNet32$\times$4   &ResNet-110 & WRN-40-2\\
Student     & ResNet8$\times$4     &VGG-8       & VGG-8       & ShuffleNetV2 &ResNet-20 & WRN-16-2 \\ \midrule
Teacher     & 79.42      &  74.64          &79.34                & 79.42 &   74.31     &75.61      \\
Vanilla Student     & 72.50      & 70.36      &70.36       & 71.82  & 69.06    &  73.26   \\ \midrule
FitNet+KD   & 74.66$\pm$0.26     & 73.22$\pm$0.21    & 73.24$\pm$0.27        & 75.15$\pm$0.19  &70.67$\pm$0.21  & 75.12$\pm$0.33  \\
FitNet+$\rm L^2$RKD & \textbf{76.36$\pm$0.21}   & \textbf{74.73$\pm$0.18}   &  \textbf{75.02$\pm$0.25}   &  \textbf{77.36$\pm$0.24}  &\textbf{71.27$\pm$0.27}  & \textbf{75.81$\pm$0.15}  \\ \midrule
AT+KD       & 74.53$\pm$0.18     & 73.48$\pm$0.19   &  74.01$\pm$0.25       & 75.39$\pm$0.29 &   70.97$\pm$0.17      & 75.32$\pm$0.15      \\
AT+$\rm L^2$RKD     & \textbf{75.47$\pm$0.17}      & \textbf{74.11$\pm$0.23}         & \textbf{75.16$\pm$0.15}  &  \textbf{76.22$\pm$0.31} &\textbf{71.23$\pm$0.21}        &\textbf{75.81$\pm$0.19}  \\ \midrule
SP+KD       & 74.02$\pm$0.24      & 73.49$\pm$0.19      & 73.52$\pm$0.25       & 74.88$\pm$0.16 &71.02$\pm$0.22     &74.98$\pm$0.28     \\
SP+$\rm L^2$RKD    & \textbf{75.35$\pm$0.27} & \textbf{73.80$\pm$0.16}     &  \textbf{74.81$\pm$0.28}     & \textbf{75.57$\pm$0.20} & \textbf{71.21$\pm$0.22}  & \textbf{75.05$\pm$0.15}            \\ \midrule
CC+KD       & 74.21$\pm$0.26   & 73.04$\pm$0.15                     & 73.48$\pm$0.29       & 74.71$\pm$0.21 &70.88$\pm$0.20           & 75.09$\pm$0.23        \\
CC+$\rm L^2$RKD     & \textbf{75.64$\pm$0.21}     & \textbf{74.69$\pm$0.24}    &  \textbf{75.17$\pm$0.21}      & \textbf{76.29$\pm$0.25} & \textbf{71.32$\pm$0.27}   & \textbf{75.69$\pm$0.28}      \\ \midrule
VID+KD      & 74.56$\pm$0.10    & 73.19$\pm$0.20        &  73.46$\pm$0.25      & 74.85$\pm$0.28  & 71.10$\pm$0.18      & 75.14$\pm$0.15    \\
VID+$\rm L^2$RKD    &\textbf{75.89$\pm$0.18}      &\textbf{74.87$\pm$0.26}     & \textbf{75.15$\pm$0.21}   & \textbf{76.55$\pm$0.27} & \textbf{71.48$\pm$0.15} & \textbf{75.68$\pm$0.25}     \\ \midrule
RKD+KD      & 73.79$\pm$0.18  &  72.97$\pm$0.08      & 73.51$\pm$0.33      & 74.55$\pm$0.23 & 70.77$\pm$0.16        & 74.89$\pm$0.20       \\
RKD+$\rm L^2$RKD   & \textbf{74.96$\pm$0.30}   & \textbf{74.58$\pm$0.15}  &  \textbf{74.75$\pm$0.29}   & \textbf{76.46$\pm$0.19} & \textbf{71.32$\pm$0.15} &\textbf{75.65$\pm$0.23}          \\ \midrule
PKT+KD      & 74.23$\pm$0.13     & 73.25$\pm$0.21  &  73.61$\pm$0.28      & 74.66$\pm$0.30  &  70.72$\pm$0.24       &75.33$\pm$0.18   \\
PKT+$\rm L^2$RKD    &\textbf{75.54$\pm$0.19}    &\textbf{74.28$\pm$0.22}  &  \textbf{75.42$\pm$0.18}      & \textbf{76.59$\pm$0.15} &\textbf{71.36$\pm$0.21}    &\textbf{75.74$\pm$0.29}   \\ \midrule
CRD+KD     & 75.46$\pm$0.25   & 74.29$\pm$0.12    &  74.58$\pm$0.27    & 76.05$\pm$0.09 & 71.56$\pm$0.16        & 75.64$\pm$0.21    \\
CRD+$\rm L^2$RKD    &  \textbf{76.51$\pm$0.20}   &\textbf{74.69$\pm$0.18}  &  \textbf{75.05$\pm$0.17}   & \textbf{76.83$\pm$0.14} & \textbf{72.00$\pm$0.27}     &\textbf{76.16$\pm$0.22}   \\ \midrule
AB+KD       & 74.40$\pm$0.27    & 73.35$\pm$0.20&   73.65$\pm$0.41      & 74.99$\pm$0.35  & 70.97$\pm$0.19     & 70.27$\pm$0.17      \\
AB+$\rm L^2$RKD     & \textbf{76.30$\pm$0.29}    &\textbf{74.67$\pm$0.19}   &  \textbf{75.24$\pm$0.20}    & \textbf{77.73$\pm$0.15}& \textbf{71.75$\pm$0.29}        &\textbf{71.51$\pm$0.26}        \\ \midrule
NST+KD      & 74.28$\pm$0.22   &  73.33$\pm$0.15    &   71.74$\pm$0.29     & 75.24$\pm$0.40  & 71.01$\pm$0.24      &74.67$\pm$0.26       \\
NST+$\rm L^2$RKD  & \textbf{75.12$\pm$0.18}    & \textbf{74.19$\pm$0.26} &\textbf{72.81$\pm$0.15}     & \textbf{76.73$\pm$0.29}    & \textbf{71.13$\pm$0.26} & \textbf{75.47$\pm$0.35} \\
 \bottomrule
\end{tabular}%
\end{table*}

\begin{table*}[!t]
\setlength{\abovecaptionskip}{0.1cm}
\setlength{\belowcaptionskip}{0.2cm}
\caption{Comparison results under few-shot settings on CIFAR-100}
\label{m4}
\centering
\begin{tabular}{lcc|cc|cc|cc}
\toprule
            & \multicolumn{2}{c|}{60\% Training Data} & \multicolumn{2}{c|}{40\% Training Data}  & \multicolumn{2}{c|}{20\% Training Data}  & \multicolumn{2}{c}{10\% Training Data}\\ \midrule
Teacher     & ResNet32$\times$4   & VGG-13    & ResNet32$\times$4  & VGG-13  & ResNet32$\times$4    & VGG-13 & ResNet32$\times$4    & VGG-13      \\
Student     & ResNet8$\times$4   & VGG-8     & ResNet8$\times$4    & VGG-8 & ResNet8$\times$4     & VGG-8& ResNet8$\times$4     & VGG-8      \\ \midrule
Teacher     & \multicolumn{1}{c}{79.42} & \multicolumn{1}{c|}{74.64} & \multicolumn{1}{c}{79.42} & \multicolumn{1}{c|}{74.64} & \multicolumn{1}{c}{79.42}  & 74.64&  \multicolumn{1}{c}{79.42}  & 74.64         \\
Vanilla Student     & 68.54        & 65.57   &  64.35            & 61.45                      & 54.70             &  52.50 &  42.76&39.30       \\ \midrule
KD          & \multicolumn{1}{c}{69.12} & \multicolumn{1}{c|}{69.90} & \multicolumn{1}{c}{66.44} & \multicolumn{1}{c|}{66.89} & \multicolumn{1}{c}{58.23}       & 59.14 & 47.95& 49.00        \\
$\rm L^2$RKD       & \multicolumn{1}{c}{\textbf{72.31}} & \multicolumn{1}{c|}{\textbf{71.62}} & \multicolumn{1}{c}{\textbf{69.55}} & \multicolumn{1}{c|}{\textbf{69.31}} & \multicolumn{1}{c}{\textbf{63.11}}       &   \textbf{63.95}& \textbf{54.56}  &\textbf{54.85}  \\ \midrule
FitNet      & \multicolumn{1}{c}{69.61} & \multicolumn{1}{c|}{66.60} & \multicolumn{1}{c}{66.97} & \multicolumn{1}{c|}{62.06} & \multicolumn{1}{c}{60.18}       & 53.57&  51.46& 39.89         \\
AT          & \multicolumn{1}{c}{69.35} & \multicolumn{1}{c|}{67.36} & \multicolumn{1}{c}{66.19} & \multicolumn{1}{c|}{65.12} & \multicolumn{1}{c}{57.72}       & 58.16& 44.70&  47.16         \\
SP          & \multicolumn{1}{c}{69.62} & \multicolumn{1}{c|}{69.76} & \multicolumn{1}{c}{65.82} & \multicolumn{1}{c|}{66.40} & \multicolumn{1}{c}{59.00}       & 58.57 & 45.44& 40.27         \\
CC          & \multicolumn{1}{c}{68.37} & \multicolumn{1}{c|}{65.37} & \multicolumn{1}{c}{64.26} & \multicolumn{1}{c|}{60.60} & \multicolumn{1}{c}{54.68}       & 51.27 & 42.74&  39.16        \\
VID         & \multicolumn{1}{c}{69.12} & \multicolumn{1}{c|}{67.29} & \multicolumn{1}{c}{65.87} & \multicolumn{1}{c|}{62.58} & \multicolumn{1}{c}{57.31}       &54.86 &  44.41&  41.07       \\
RKD         & \multicolumn{1}{c}{67.71} & \multicolumn{1}{c|}{66.18} & \multicolumn{1}{c}{63.51} & \multicolumn{1}{c|}{62.32} & \multicolumn{1}{c}{52.29}       &52.13 & 39.19 &39.70        \\
PKT         & \multicolumn{1}{c}{70.48} & \multicolumn{1}{c|}{69.21} & \multicolumn{1}{c}{66.41} & \multicolumn{1}{c|}{65.97} & \multicolumn{1}{c}{59.06}       & 58.08 & 43.50  &41.15        \\
CRD         & \multicolumn{1}{c}{71.29} & \multicolumn{1}{c|}{70.46} & \multicolumn{1}{c}{68.15} & \multicolumn{1}{c|}{66.27} & \multicolumn{1}{c}{59.38}       &57.57   & 48.23&  46.33       \\
AB          & \multicolumn{1}{c}{69.25} & \multicolumn{1}{c|}{65.98} & \multicolumn{1}{c}{65.30} & \multicolumn{1}{c|}{63.07} & \multicolumn{1}{c}{58.48}  & 56.55  & 48.61& 48.27    \\
FT          & \multicolumn{1}{c}{67.05} & \multicolumn{1}{c|}{64.88} & \multicolumn{1}{c}{63.38} & \multicolumn{1}{c|}{60.37} & \multicolumn{1}{c}{53.85}  &50.42 &   39.55& 38.10        \\
NST         & \multicolumn{1}{c}{69.87} & \multicolumn{1}{c|}{67.12} & \multicolumn{1}{c}{66.24} & \multicolumn{1}{c|}{63.56} & \multicolumn{1}{c}{60.27}       & 56.63 &51.91   &  47.44         \\ \bottomrule
\end{tabular}%
\end{table*}

\begin{table*}[!t]
\setlength{\abovecaptionskip}{0.1cm}
\setlength{\belowcaptionskip}{0.2cm}
\caption{Compatibility performances under few-shot settings on CIFAR-100}
\label{comfew}
\centering
\begin{tabular}{lcc|cc|cc|cc}
\toprule
            & \multicolumn{2}{c|}{60\% Training Data} & \multicolumn{2}{c|}{40\% Training Data}  & \multicolumn{2}{c|}{20\% Training Data}  & \multicolumn{2}{c}{10\% Training Data}\\ \midrule
Teacher     & ResNet32$\times$4   & VGG-13    & ResNet32$\times$4  & VGG-13  & ResNet32$\times$4    & VGG-13 & ResNet32$\times$4    & VGG-13      \\
Student     & ResNet8$\times$4   & VGG-8     & ResNet8$\times$4    & VGG-8 & ResNet8$\times$4     & VGG-8& ResNet8$\times$4     & VGG-8      \\ \midrule
FitNet+KD   &  72.09    &  69.21        & 69.64 & 65.94   &   63.36 &  58.41& 54.72 & 46.30        \\
FitNet+$\rm L^2$RKD  & \textbf{74.03}    &   \textbf{71.87}  &  \textbf{71.80} &\textbf{69.68} &   \textbf{66.58}  &  \textbf{63.57}& \textbf{58.79}&\textbf{50.56}     \\ \midrule
AT+KD       &  71.54    & 70.61  &   68.01 &   68.22  &     60.89      & 62.78& 50.22 & 53.91         \\
AT+$\rm L^2$RKD      & \textbf{72.59}  &   \textbf{72.20}  & \textbf{70.17}  & \textbf{69.60}&   \textbf{64.21}  & \textbf{64.50}&\textbf{55.60} &\textbf{55.89}   \\ \midrule
SP+KD       &   70.45 & 69.70  &  67.22 & 66.73 &60.53               &    60.35& 49.94& 45.62       \\
SP+$\rm L^2$RKD      &  \textbf{72.41} & \textbf{71.67} & \textbf{69.33} &  \textbf{69.65}   &   \textbf{63.90}   &\textbf{63.56}&\textbf{54.41}&\textbf{50.97}        \\ \midrule
CC+KD       &   70.67 &   69.38     & 66.81& 66.27 &       59.03      &   58.60 & 48.54  & 48.37     \\
CC+$\rm L^2$RKD      &  \textbf{72.36}& \textbf{71.66}  &\textbf{69.40} & \textbf{69.35}  &  \textbf{63.50}       &  \textbf{63.33}&\textbf{53.51}& \textbf{53.15} \\ \midrule
VID+KD      &   70.00 &69.54 & 67.65&  66.42 &   60.03   & 58.68& 47.84&   46.81        \\
VID+$\rm L^2$RKD     &\textbf{72.75}  & \textbf{72.41}  & \textbf{70.23} & \textbf{69.75} &    \textbf{64.49}  &   \textbf{64.18}&\textbf{54.56} &\textbf{54.54}        \\ \midrule
RKD+KD      &   70.33  & 69.74  &  66.63      &66.15  &       58.54                     &   58.95& 46.43&  48.20         \\
RKD+$\rm L^2$RKD     & \textbf{72.16}  &  \textbf{71.69} &   \textbf{69.24} &  \textbf{69.60}  &  \textbf{63.08}    &  \textbf{63.28}& \textbf{51.77}&\textbf{56.62}  \\ \midrule
PKT+KD      &  70.98 &   70.18    & 67.43    & 66.13  &    60.26   &   59.48  & 48.78& 47.88      \\
PKT+$\rm L^2$RKD     & \textbf{72.43}& \textbf{72.21}   &\textbf{69.81}&  \textbf{69.51}&  \textbf{63.90} &\textbf{63.46}& \textbf{53.29}&\textbf{53.87}  \\ \midrule
CRD+KD      & 71.97&  70.74   &68.83  & 66.84   &61.27 &  59.07& 48.58  &47.94         \\
CRD+$\rm L^2$RKD     &\textbf{72.82}&  \textbf{71.51} &  \textbf{69.43}  & \textbf{68.91}& \textbf{63.85} & \textbf{62.58} & \textbf{52.95} &\textbf{53.82}          \\ \midrule
AB+KD       &   70.04 &69.95 & 67.75&67.47 & 61.74 &  62.69 &  53.76& 60.69       \\
AB+$\rm L^2$RKD      & \textbf{73.36}  &  \textbf{72.78} & \textbf{70.80}&  \textbf{71.47}  &  \textbf{65.58}& \textbf{68.70}& \textbf{59.41}&\textbf{64.07}  \\ \midrule
NST+KD      &  71.15& 69.36  &  67.92   &66.87    &     62.37  &  61.04& 55.05& 52.49         \\
NST+$\rm L^2$RKD     &  \textbf{72.27}&   \textbf{71.84}  &  \textbf{69.66}   & \textbf{69.68}  &   \textbf{65.33}  &\textbf{65.01} & \textbf{58.83}&\textbf{58.22}   \\ \bottomrule
\end{tabular}%
\end{table*}

\begin{table}[!t]
\centering
\setlength{\abovecaptionskip}{0.1cm}
\setlength{\belowcaptionskip}{0.1cm}
\caption{S-T DIFs (shape differences) on CIFAR-100}
\label{m_dif}
\begin{tabular}{lcccccccc}
\toprule
Teacher        & ResNet32$\times$4       & WRN-40-2 & WRN-40-2  & VGG-13  \\
Student         & ResNet8$\times$4        & WRN-16-2   & WRN-40-1   & VGG-8   \\ \midrule %
KD               & 2.81                     & 2.74         & 2.94    &    1.77    \\ %
$\rm L^2$RKD & \textbf{1.59}       & \textbf{2.11} & \textbf{2.21}  & \textbf{1.25} \\ \bottomrule %
\end{tabular}%
\vskip -0.0in
\end{table}


\subsection{Performances under Few-shot Scenario}
In many practical situations, it can happen that a powerful model is released, but only a few data samples are publicly accessible due to the privacy or confidentiality issues in various domains such as medical and industrial domains.
It is essential for knowledge distillation approaches to work in these data-limited cases.
We evaluate $\rm L^2$RKD in the few-shot scenario where knowledge is transferred from a powerful teacher to a student with limited training data available.

\subsubsection{Comparison Results under Few-shot Scenario}
Table \ref{m4} presents the performances of different approaches under few-shot settings.
We observe that $\rm L^2$RKD substantially outperforms KD and the other approaches in all the cases with 60\%, 40\%, 20\%, and 10\% training data available, which demonstrates the superiority of $\rm L^2$RKD under few-shot settings.
This also indicates the importance and usefulness of the knowledge in local, linear regions on data-limited situations.
We also notice that the superiority of $\rm L^2$RKD becomes more obvious when the data is sparser.
For example, when 60\% training data is available, the absolute accuracy improvement of $\rm L^2$RKD over KD on teacher ResNet32$\times$4 and student ResNet8$\times$4 is 3.19\% (i.e., 72.31\% - 69.12\%).
However, when 10\% training data is available, the absolute accuracy improvement is 6.61\% (i.e., 54.56\% - 47.95\%).
The improvement of $\rm L^2$RKD is larger when the data becomes sparser.
This reason is that when the data is highly sparse, the argument holds strongly that fitting sparse data points cannot enable the students to well capture the local shape of the teacher.
$\rm L^2$RKD addresses this issue by enforcing the student to mimic the behavior of the teacher in local regions, thus making the student better capture the local shape of the teacher and achieving a better performance.

\subsubsection{Compatibility with Existing Approaches under Few-shot Scenario}
We also evaluate the compatibility of $\rm L^2$RKD with the existing approaches under few-shot Scenario.
The compatibility performances are reported in Table \ref{comfew}.
It is observed that these approaches achieve much better performances when combined with $\rm L^2$RKD than when combined with KD, which demonstrates the superior compatibility of $\rm L^2$RKD with the existing methods in the data-limited cases.
This also indicates that the knowledge in local, linear regions is outstandingly important in data-limited situations.

\subsection{Shape Differences}
\label{shape}
In this part, we verify that the students trained with $\rm L^2$RKD better capture the local shapes of the teachers than those trained with KD.
The local shape of a function can be represented by a set of pairs $(x, y)$ where $x$ is the input and $y$ is the output of the function.
To measure the shape difference, we report the average mean square \textbf{student-teacher output logit differences (S-T DIFs)} by using test data as inputs. 
We adopt four pairs of students and teachers on CIFAR-100 and report the shape differences in Table \ref{m_dif}.
It is observed that S-T DIFs of $\rm L^2$RKD are consistently smaller than those of KD, which demonstrates that the student shapes of $\rm L^2$RKD are closer to the teacher shapes and indicates that the knowledge in locally linear regions is beneficial for the students to capture the local shapes of the teachers.

\begin{table}[!h]
\caption{Comparison results of $\rm L^2$RKD with NoiseKD on CIFAR-100}
\label{n}
\centering
\begin{tabular}{lccc}
\toprule
Teacher           & VGG-13                          & WRN-40-2                        & ResNet32$\times$4                \\
Student           & VGG-8                           & WRN-40-1                        & ResNet8$\times$4                 \\ \midrule
Teacher           & 74.64                           & 75.61                           & 79.42                           \\
Vanilla Student   & 70.36                           & 71.98                           & 72.50                           \\\midrule
KD                & 72.98                           & 73.54                           & 73.33                           \\
NoiseKD-$\mathcal{N}(0, 0.1)$   & 69.58                           & 68.26                           & 66.75                           \\
NoiseKD-$\mathcal{N}(0, 0.05)$   & 72.67                           & 72.20                           & 71.76                           \\
NoiseKD-$\mathcal{N}(0, 0.01)$   & 73.26                           & 73.58                           & 73.70                           \\
NoiseKD-$\mathcal{N}(0, 0.005)$  & 72.94                           & 73.47                           & 72.96                           \\
$\rm L^2$RKD      & $\textbf{74.54}$ &  $\textbf{75.38}$ & \textbf{75.67} \\ \bottomrule
\end{tabular}
\end{table}
\subsection{Comparison with Noise Region Distillation}
$\rm L^2$RKD explores knowledge of a teacher by distilling on in local, linear regions.
Injecting small, random noise to training data points can also form many local regions around each training data point.
These local regions are determined by the type of the noise.
Intuitively, distilling on these regions can also explore more knowledge in the teacher.
We call this method NoiseKD.
We compare $\rm L^2$RKD with NoiseKD.
We inject the commonly used Gaussian nosie to the training data points.
We grid search the best hyperparameter for NoiseKD by using different levels of Gaussian noise, i.e., $\mathcal{N}(0, 0.1)$, $\mathcal{N}(0, 0.05)$, $\mathcal{N}(0, 0.01)$, and $\mathcal{N}(0, 0.005)$.
Table \ref{n} reports the comparison results.
It is observed that when the noise in NoiseKD is large (e.g., $\mathcal{N}(0, 0.1)$ and $\mathcal{N}(0, 0.05)$), NoiseKD even underperforms KD, which indicates that large noise is harmful to knowledge distillation.
When noise is relatively small (e.g., $\mathcal{N}(0, 0.01)$), NoiseKD slightly improves the performances over KD, which indicates that small noise is useful for knowledge distillation.
We also see that $\rm L^2$RKD consistently outperforms NoiseKD with different levels of noise, which demonstrates the superiority of $\rm L^2$RKD.

\subsection{Comparison with Generative Network Distillation}
As observed from the error bound $O(\sqrt{\frac{C}{n}})$ in Section \ref{conver}, increasing the number of unlabeled samples for distillation can reduce the fitting error and thus improves the student perform.
One natural idea is to use generative adversarial networks (GANs) \cite{goodfellow2014generative, arjovsky2017wasserstein, liu2018teacher} to learn the data distribution and then use the generator in GANs to generate fake unlabeled data samples for distillation.
We denote this method by GAN-KD.
However, there are two issues in GAN-KD: first, training GANs is computationally expensive especially for large datasets (e.g., ImageNet) while $\rm L^2$RKD uses freely obtained in-distribution and out-of-distribution points in local, linear regions; second, the diversity and quality of the generated fake data from GANs are highly limited by training data samples. In other words, GANs cannot accurately learn the real data sample distribution $P(x)$, just like we cannot obtain 100\% test accuracy by training a DNN classifier on the training data samples and their ground truth of CIFAR-100.
The difficulty of using GANs to learn the real data sample distribution $P(x)$ from training data points is as the same as that of using a DNN to learn the conditional distribution $P(y|x)$ from training data points and their ground truth.\par

We compare $\rm L^2$RKD with GAN-KD.
We first use a GAN to learn the data sample distribution of CIFAR-10 \cite{krizhevsky2009learning} as GANs can easily converge on CIFAR-10.
Note that infinite faker samples can be generated by the GAN, but their qualities and diversities are highly limited by the training data.
Thus, we grid search the optimal ratio of the number of the training samples to the number of the generated faker samples for distillation.
We find that GAN-KD achieves the best performance at ration 1.
The comparison results are reported in Table \ref{cgan}.
It is observed that GAN-KD improves the performances over KD, but it underperforms $\rm L^2$RKD substantially.
This indicates that GANs can generate some useful fake samples for knowledge distillation and also verifies that the diversity and usefulness of these samples are highly constrained by the training data.

\begin{table}[!t]
\caption{Comparison results of $\rm L^2$RKD with GAN-KD on CIFAR-10}
\label{cgan}
\centering
\begin{tabular}{lcccc}
\toprule
Teacher                 & WRN-40-2                & VGG-13                      \\
Student                   & WRN-16-1               & MobileNetV2                                \\ \midrule
Teacher                  & 94.80               & 94.01                                             \\
Vanilla Student           & 91.32              & 89.19                                                \\\midrule
KD                     &91.77                  & 89.21                                             \\
GAN-KD           &      92.23      &          89.25                                                 \\
$\rm L^2$RKD           &     \textbf{93.11}     & \textbf{ 90.52   }                                  \\  \bottomrule
\end{tabular}
\end{table}


\section{Conclusion and Future Work}
We observe that only distilling knowledge at sparse data points cannot enable a student to well capture the local shape of a teacher function.
To address this issue, we have proposed $\rm L^2$RKD which transfers the knowledge in local, linear regions from a teacher to a student.
This is achieved by enforcing the student to fit the teacher in local, linear regions.
To the end, the student can better capture the local shape of the teacher and thus obtains a better performance.
We compare $\rm L^2$RKD with more than 10 state-of-the-art approaches on four benchmark datasets with varieties of modern network architecture.
Despite its simplicity, $\rm L^2$RKD beats KD and the other state-of-the-art approaches substantially, shows superiority under few-shot settings, and is compatible with the existing approaches to further improve their performances significantly.
\par

$\rm L^2$RKD addresses knowledge distillation from a novel data perspective.
The existing work has not sufficiently explored the knowledge in the teacher function. 
$\rm L^2$RKD addresses this issue by distilling on local, linear regions.
These local, linear regions are freely obtained by using linear algebra without training any additional generative models.
The simple nature of $\rm L^2$RKD is a beauty and of much practical value, i.e., easy to understand and implement yet effective.
However, local, linear regions may not be the optimal solution to this problem.
In the future, we will try to automatically learn the optimal region for exploring the knowledge in the teacher.

\appendix[Derivation From (\ref{3}) to (\ref{fit})]

We provide the details for the derivation from ($\ref{3}$) to ($\ref{fit}$).\par

($\ref{3}$) can be expanded as:

\begin{equation}
\label{p1}
\begin{aligned}
\mathcal{L}_{KL}\left(f_S, f_T, x\right) = \tau^2 \sigma\left(\frac{f_T\left(x\right)}{\tau}\right)log \sigma\left(\frac{f_T\left(x\right)}{\tau}\right) \\ -\tau^2 \sigma\left(\frac{f_T\left(x\right)}{\tau}\right)log \sigma\left(\frac{f_S\left(x\right)}{\tau}\right)
\end{aligned}
\end{equation}
The first term in the right hand side of (\ref{p1}) is a constant.
Then the gradient of $\mathcal{L}_{KL}$ with respect to each logit of $f_S(x)$ is written as:
\begin{equation}
\label{p5}
\frac{\partial{\mathcal{L}_{KL}}}{\partial{f_S(x)}[i]} = \tau\left( \sigma\left(\frac{f_S\left(x\right)[i]}{\tau}\right) - \sigma\left(\frac{f_T\left(x\right)[i]}{\tau}\right)\right)
\end{equation}
where $f_S(x)[i]$ is the $i$th logit in $f_S(x)$.
(\ref{e4}) is equivalent to (\ref{p5}) with multiple logits.
Expanding (\ref{p5}) leads to the following:
\begin{equation}
\label{p6}
\frac{\partial{\mathcal{L}_{KL})}}{\partial{f_S\left(x\right)[i]}} = \tau \left( \frac{exp\left(\frac{f_S\left(x\right)[i]}{\tau}\right)} { \sum_j exp\left(\frac{f_S\left(x\right)[j]}{\tau} \right)}   - 
\frac{exp\left(\frac{f_T\left(x\right)[i]}{\tau}\right)} { \sum_j exp\left(\frac{f_T\left(x\right)[j]}{\tau} \right)}
\right)
\end{equation}

The Taylor Series for $exp(x)$ is:
\begin{equation}
\label{p7}
 exp(x)=1+x+\frac{x^2}{2!}+\frac{x^3}{3!}+, ..., +\frac{x^n}{n!}
\end{equation}
where $!$ is the factorial function.

When $\tau$ is large compared with the magnitude of the logits, (\ref{p6}) can be approximated as the following:
\begin{equation}
\label{p8}
\frac{\partial{\mathcal{L}_{KL}}}{\partial{f_S\left(x\right)[i]}} \approx  \tau \left( \frac{1+\frac{f_S\left(x\right)[i]}{\tau}} { K+ \sum_j \frac{f_S\left(x\right)[j]}{\tau} }   - 
\frac{1+\frac{f_T\left(x\right)[i]}{\tau}}{ K+\sum_j \frac{f_T\left(x\right)[j]}{\tau} } \right)
\end{equation}
where $K$ is the total number of classes.
(\ref{e5}) is equivalent to (\ref{p8}) with multiple logits.
With the assumption that the logits of a DNN have mean zero, we can obtain: 
\begin{equation}
\label{p9}
\frac{\partial{\mathcal{L}_{KL}}}{\partial{f_S(x)[i]}} \approx  \frac{1}{K}\left(f_S\left(x\right)[i]-f_T\left(x\right)\right[i])
\end{equation}
Finally, we obtain (\ref{fit}) that is equivalent to (\ref{p9}) with multiple logits:
\begin{equation}
\label{p10}
\frac{\partial{\mathcal{L}_{KL}}}{\partial{f_S(x)}} \approx  \frac{1}{K}\left(f_S\left(x\right)-f_T\left(x\right)\right)
\end{equation}

%





\ifCLASSOPTIONcaptionsoff
  \newpage
\fi



\bibliographystyle{IEEEtran}
\bibliography{LRKD}
%

%

\end{document}